\documentclass[10pt,twocolumn,letterpaper]{article}

\usepackage{iccv}
\usepackage{times}
\usepackage{epsfig}
\usepackage{graphicx}
\usepackage{amsmath}
\usepackage{amssymb}
\usepackage{booktabs}
\usepackage[affil-it]{authblk}

\usepackage[pagebackref=true,breaklinks=true,letterpaper=true,colorlinks,bookmarks=false]{hyperref}

\iccvfinalcopy 

\def\iccvPaperID{SG2RL-26} 

\ificcvfinal\fi

\begin{document}

\title{Dynamic Scene Graph Representation for Surgical Video}
\author[1, 2]{Felix Holm \thanks{Corresponding Author: felix.holm@tum.de}\thanks{These authors contributed equally to this work.}}
\author[2]{Ghazal Ghazaei\footnote[2]}
\author[1]{Tobias Czempiel}
\author[1]{Ege Özsoy}
\author[3]{Stefan Saur}
\author[1]{Nassir Navab}

\affil[1]{Chair for Computer Aided Medical Procedures, Technical University Munich, Germany}
\affil[2]{Carl Zeiss AG, Munich, Germany}
\affil[3]{Carl Zeiss Meditec AG, Oberkochen, Germany}


\maketitle
\ificcvfinal\thispagestyle{empty}\fi

\begin{abstract}
   Surgical videos captured from microscopic or endoscopic imaging devices are rich but complex sources of information, depicting different tools and anatomical structures utilized during an extended amount of time. Despite containing crucial workflow information and being commonly recorded in many procedures, usage of surgical videos for automated surgical workflow understanding is still limited. 
    In this work, we exploit scene graphs as a more holistic, semantically meaningful and human-readable way to represent surgical videos while encoding all anatomical structures, tools, and their interactions. To properly evaluate the impact of our solutions, we create a scene graph dataset from semantic segmentations from the CaDIS and CATARACTS datasets. We demonstrate that scene graphs can be leveraged through the use of graph convolutional networks (GCNs) to tackle surgical downstream tasks such as surgical workflow recognition with competitive performance. Moreover, we demonstrate the benefits of surgical scene graphs regarding the explainability and robustness of model decisions, which are crucial in the clinical setting. 
\end{abstract}

\section{Introduction}
\textbf{Videos for Surgical Workflow Understanding} capture the most integral parts of surgeries from global surgery information to the most atomic actions and tasks. Automated surgical workflow understanding solutions rely on such rich, dense, and complex sources of information to extract useful surgical knowledge such as surgical phases, steps, events as well as instruments and anatomical structures. Surgical Workflow analysis could further promote assisting the surgeons via automated report generation, intra-operative warnings, robotic assistance, workflow optimization, and education.

Accordingly, the field of surgical data science has seen a significant amount of research in the recognition of the surgical workflow, enabled by the introduction of publicly available datasets, such as Cholec80 for laparoscopic cholecystectomy surgeries~\cite{twinanda_endonet_2016} or CATARACTS for microscopic cataracts surgery~\cite{al2019cataracts}. Methods applied in this field range from Convolutional Neural Networks (CNN)~\cite{twinanda_endonet_2016} to extract latent representations of visual image features to various temporal modeling methods such as Long Short-Term Memory (LSTM)~\cite{Garrow2020}, Temporal Convolutions \cite{martel_tecno_2020}, or transformer-based methods~\cite{czempiel_opera_2021,DBLP:journals/corr/abs-2103-09712}. These methods were very successful in steadily improving on the task of surgical workflow recognition, their real-world application in the clinical setting is however still strongly limited. Further advancement in the field of surgical data science is striving for a more holistic and robust understanding of surgery such that tools and anatomies are explicitly recognized and their interactions over time can be meticulously analyzed. 





\textbf{Scene Graphs} \cite{DBLP:journals/corr/KrishnaZGJHKCKL16,DBLP:journals/corr/XuZCF17} have been proposed to represent scenes with multiple actors, objects, and relationships between them in a graphical way. Relationships between entities are visualized as edges in a graph where each entity corresponds to a node. Spatio-temporal scene graphs~\cite{DBLP:journals/corr/abs-1912-06992} have been used specifically to analyze and differentiate relationships between objects over time. The scene graph representation is a powerful modeling technique to encode a scene on a fine granular level in a human-understandable way and creates the foundation for different complex downstream tasks such as visual question answering or image captioning. As an abstract, but still human-readable representation of a scene, scene graphs also offer benefits in terms of explainability and reasoning, which is highly desirable, especially in the medical domain.
Scene graphs have previously been introduced in the medical domain for a holistic understanding of the OR, where the surgical staff and their interactions are modeled~\cite{ozsoy_4d-or_2022}.  For the 'internal view' of the surgery, such as from an endoscope or surgical microscope, surgical action triplets for cholecystectomy~\cite{sharma_rendezvous_2022} and scene graphs for robotic nephrectomy~\cite{islam_learning_2020} have been proposed. 
These methods model the interactions between surgical tools and single target anatomies as graphs but do however not model the full anatomical structure and the relations between them.

In this work, we aim to investigate the utility of surgical scene graphs as a representation of surgical microscopic video for surgical workflow understanding. We utilize scene graphs to represent the complete scene, including all visible anatomies, tools, and their interactions. To avoid the cost of detailed annotations for these experiments, we make use of the CATARACTS and CaDIS datasets. The CATARACTS dataset contains microscopic cataract surgery videos and their corresponding surgical phase annotations to evaluate surgical workflow understanding methods. The CaDIS dataset is a subset of CATARACTS containing semantic segmentation annotations, which allows us to leverage off-the-shelf state-of-the-art semantic segmentation models~\cite{pissas2021effective,cheng2022masked} to localize surgical scene components and further extract their relations to construct a scene graph. The extracted scene graphs are then exploited within a Graph Convolutional-based framework to perform the task of surgical phase recognition. As surgical phases are best presented considering the temporal inter-relations between the frames, we further reinforce our solution with temporal connections among the neighboring static scene graphs to construct a Dynamic Scene Graph (DSG). We highlight the performance of our method and further investigate the benefits of scene graphs in terms of robustness and explainability of predictions.

The features of our suggested solution are as follows: 

\begin{enumerate}
    \item Leveraging off-the-shelf semantic segmentation solutions and available surgical video datasets for creation of a scene graph dataset 
    \item Implementation of a GCN-based framework to understand and analyze surgical videos via a minimal yet holistic intermediate representation incorporating both spatial and temporal information within surgeries
    \item Exploiting the suggested representation on surgical downstream task of phase segmentation to highlight the benefits namely usability, robustness and explainability. 
\end{enumerate}


\begin{figure*}
    \centering
    \includegraphics[width=\textwidth,height=0.5\textwidth]{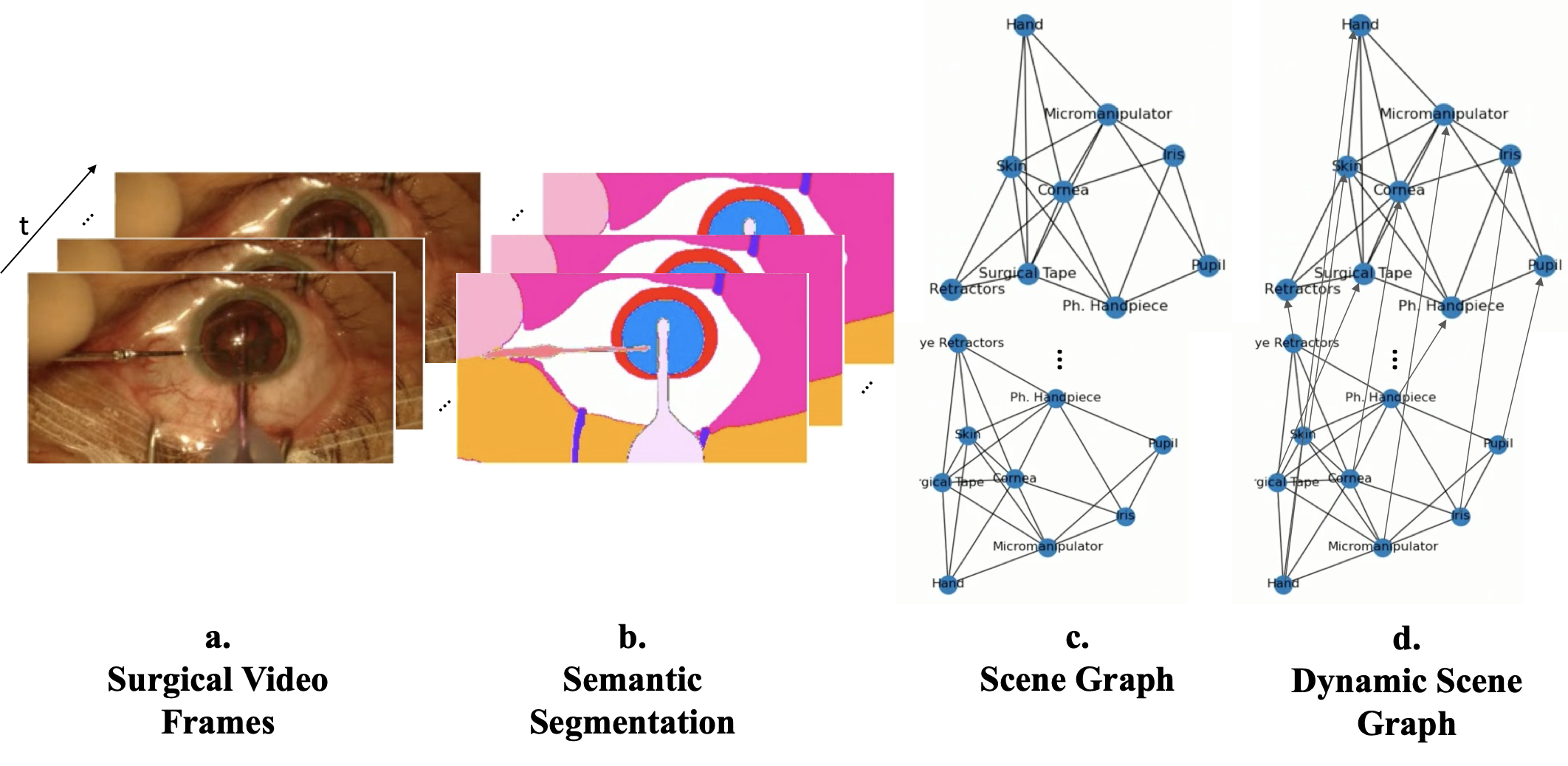}
    \caption{Illustration of surgical Dynamic Scene Graph (DSG) creation: a) Video recorded from the microscopic view of cataract surgery b) semantic segmentation maps indicating anatomical structures and tools c) static scene graphs constructed for each frame highlighting anatomies, tools and their relations d) dynamic scene graph for a sequence of frames indicating the tools, anatomies and their relations across both space (within each frame) and time}
    \label{fig:intro_figure}
\end{figure*}

\section{Methods}

\subsection{Semantic Segmentation}
We build upon the state-of-the-art solutions in semantic scene segmentation for the CATARACTS dataset, namely OCRNet~\cite{pissas2021effective} and Mask2Former~\cite{cheng2022masked}, to precisely localize surgical scene components.  
\textbf{OCRNet} creates a soft segmentation map by dividing the contextual pixels into a set of soft object regions each assigned to a class. The representations of pixels in an object region are then aggregated to predict the representation of that object region. Finally, each pixel representation is augmented with the object-contextual-representation (OCR) which is a weighted aggregation of all object region representations.
\textbf{Mask2Former} enhances the extracted feature maps from a backbone ConvNet using a pixel decoder module which outputs high-resolution features. The features are passed into a transformer decoder, creating a set of binary mask and class predictions conditioned on the pixel decoder's features.

\subsection{Graph Construction}

\textbf{Static Graph}
Given an input RGB frame, the segmentation model generates a global 2D segmentation map as visualized in Figure~\ref{fig:intro_figure}.b, which is composed of a set of segments $S=\{s_1, ..., s_n\}$. To generate the scene graph, each segment is represented as a node $N_i$ of the graph $G = (N, E)$ where $N=\{N_1, ..., N_n\}$ and $E$ represents the edges between two nodes. 
Each node contains a set of features $X \in {R} ^{N\times d}$. We incorporated a variety of features into the scene graphs to study their individual impact. The baseline node feature set only encodes the segment classification from the segmentation model. Additionally, we encode the spatial position of the segment, the relative size of the segment and, in the case of Mask2Former, the instance embedding of each segment. We provide ablation experiments on the features used for graph construction in section~\ref{sec:results}.
The undirected edges of the static graph are defined based on the geometrical relation between each two segments $(s_i,s_j)$. If the two segments are connected in the segmentation mask, their corresponding nodes are also connected.


\textbf{Dynamic Scene Graph (DSG)}
To leverage temporal relations to create a more holistic spatio-temporal understanding, we aggregate multiple Static Graphs as defined above and connect each node $N_i$ to the nodes belonging to the same class from adjacent temporal steps. The DSG therefore consists of the nodes $N_{t_i}$ and edges $E_{t_i}$ of the static graph of each timestep and the additional temporal connections $E_{t_i\xrightarrow{}t_{i+1}}$, resulting in $G_{DSG} = (N_{t_0} + N_{t_1} + \ldots , E_{t_0} + E_{t_0\xrightarrow{}t_{1}} + E_{t_1} + \ldots)$. This allows the changes of each segment $s_i$ over time to be captured and analyzed along the added temporal connections. We vary the number of static graphs that are aggregated to form the dynamic graph, which we refer to as 'Window'. We also consider 'Dilation', where we use static graphs with larger temporal distances to form the dynamic graph to increase the covered temporal context while limiting the graph size. We investigate the impact of these parameters in section~\ref{sec:results}.
For dynamic graphs, we also add an encoding of the relative temporal position of each node to the node features, which we also ablate in section~\ref{sec:results}.

\subsection{Graph Convolutional Network} 
\begin{figure*} [h]
    \centering
    \includegraphics[width=\textwidth]{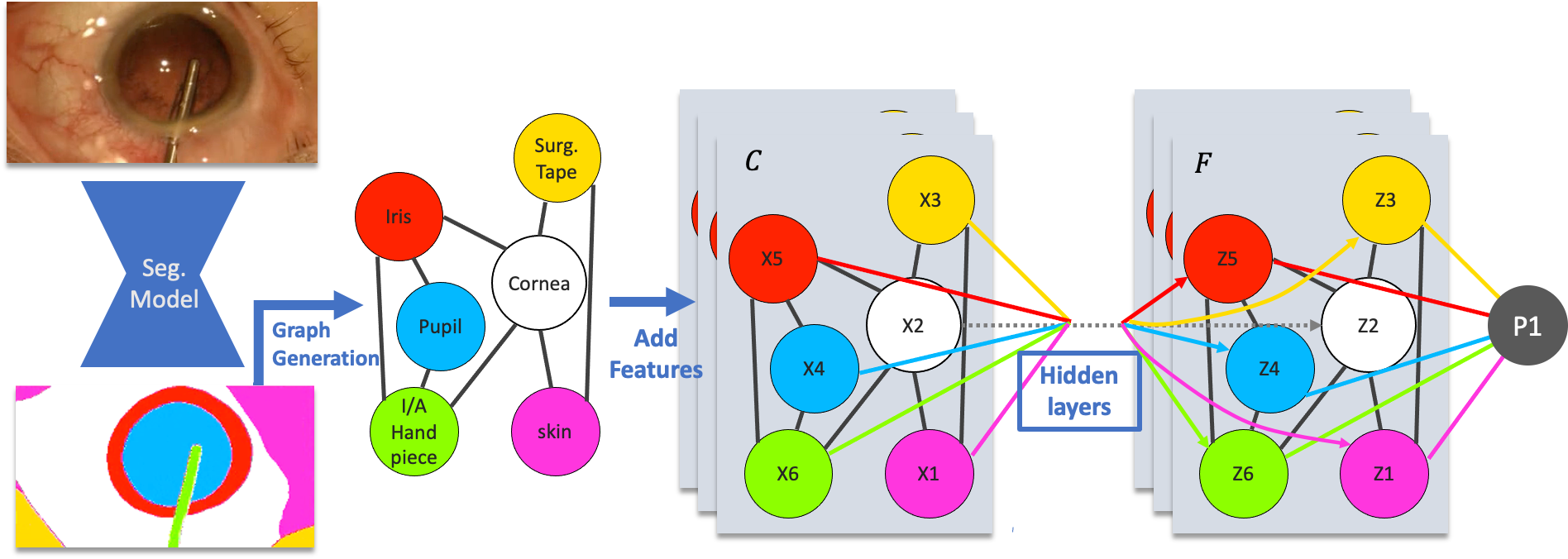}
    \caption{Overview of the proposed pipeline: Video frames are segmented based on their semantic classes $s_i$ which can then be leveraged for construction of a scene graph with each semantic class as a node $N_i$. Each node will be connected to another if they are adjacent to each other in the segmentation map. The nodes can be further reinforced with custom features. The scene graph is then passed into a multi-layer GCN trained on the downstream task of surgical phase recognition. $C$ and $F$ indicate the number of input and feature maps respectively.}
    \label{fig:method_figure}
\end{figure*}

To solve the downstream task of phase recognition, we design a multi-layer graph convolutional network (GCN)~\cite{kipf2016semi}, taking the scene graph as input.
The multi-layer GCN is followed by a global add-pooling layer that aggregates the features from all nodes, followed by a fully connected layer and $\text{Softmax}(x_{i}) = \frac{\exp(x_i)}{\sum_j \exp(x_j)}$ to predict the probabilities for each phase class.
We use the cross-entropy loss as our objective function on the labeled frames of videos to optimize the model parameters. Figure~\ref{fig:method_figure} shows the overall model structure.

\section{Evaluation}

\subsection{Datasets} 
\textbf{CATARACTS}~\cite{ac97-8m18-21,al2019cataracts} consists of 50 cataract surgery videos with a resolution of $1920\times1080$ pixels at $30$ frames per second (fps) performed at Brest University Hospital. The average duration of Cataract surgeries within this dataset is $10$ minutes and $56$~s. The dataset as published for the CATARACTS2020 challenge defines a train set (25 videos), val set (5 videos), and test set (20 videos). Each frame of the videos is annotated with surgical phases by a medical doctor and an ophthalmology nurse. There is a total of 18 phases (excluding the idle phase): 1)~Toric Marking, 2)~Implant Ejection, 3)~Incision, 4)~Viscodilatation, 5)~Capsulorhexis, 6)~Hydrodissetion, 7)~Nucleus Breaking, 8)~Phacoemulsification, 9)~Vitrectomy, 10)~Irrigation/Aspiration, 11)~Preparing Implant, 12)~Manual Aspiration, 13)~Implantation, 14)~Positioning, 15)~OVD Aspiration 16)~Suturing, 17)~Sealing Control, 18)~Wound Hydratation.

\textbf{CaDIS}~\cite{grammatikopoulou2019cadis} is a subset of CATARACTS including a total of $4670$ pixel-wise annotated images. There are three increasingly granular semantic segmentation tasks defined on CaDIS, from which we take Task II, semantically representing sufficient information for the downstream task of phase segmentation. Task II consists of $17$ classes: 0)~Pupil 1)~Surgical Tape, 2)~Hand, 3)~Eye Retractors, 4)~Iris, 5)~Skin, 6)~Cornea, 7)~Cannula, 8)~Capsulorhexis Cystotome, 9)~Tissue Forceps, 10)~Primary Knife, 11)~Phaco Handpiece, 12)~Lens Injector, 13)~I/A Handpiece, 14)~Secondary Knife, 15)~Micromanipulator, 16)~Capsulorhexis Forceps.

\subsection{Implementation details}
\textbf{Segmentation Models}
We use the provided pre-trained weights for the OCRNet model~\cite{pissas2021effective}, while we trained the Mask2Former model with the Swin small backbone on the CaDIS dataset using the implementation provided by~\cite{mask2formercode}.
The segmentation models are evaluated on the full CATARACTS dataset (at $1$ fps) to create semantic segmentation maps, out of which static scene graphs are generated.

\textbf{Graph Construction}
The scene graphs are constructed and populated with features extracted from the segmentation depending on the experiment. The sizes of these feature vectors are: 15 (segment classification), 16 (spatial position), 1 (segment size), 16 (temporal position),  100 (M2F class query embedding).

\textbf{Graph Convolutional Network}
Our scene graph-based phase segmentation model consists of 8 graph convolution layers with hidden dimensions of 64, 64, 128, 128, 192, 128, 64, 64, and a fully convolutional layer. Since the model and constructed scene graph dataset are relatively small, the GCN can be trained without GPU hardware acceleration. This highlights another strength of scene graphs as a lightweight video representation.

 
\subsection{Results and Discussion}
\label{sec:results}
\paragraph{Semantic Segmentation}

Table~\ref{tab:semantic_seg} highlights the results of Mask2Former compared with state of the art, OCRNet. Mask2Former performs better mean Intersection over Union (mIoU) overall and more specifically indicates prior performance for localization and segmentation of instruments which is crucial for the downstream task of phase segmentation. It is worth to note that both models still present noisy pixelwise segmentation and misclassification for scene representations that are more distinct from the majority of dataset distribution.

\begin{table}[]
    \centering
    \caption{Semantic segmentation results of used models on Task~II of CaDIS~\cite{grammatikopoulou2019cadis} test set}
    \begin{tabular}{lccc}
         &  OCRNet~\cite{pissas2021effective} &  Mask2Former~\cite{cheng2022masked} \\
         \toprule
         mIoU (Anatomy) &\textbf{90.63} & 83.14 \\
         mIoU (Instruments) &76.89 & \textbf{83.58} \\
         \midrule
         mIoU (All classes) &79.09 & \textbf{80.97} \\
         \bottomrule
    \end{tabular} \label{tab:semantic_seg}
\end{table}

\paragraph{Graph Representation}
Table~\ref{tab:ablation_architecture} illustrates the impact of features on our static scene graph within extensive ablation studies, from only having class embedding to enriching the graph with temporal and positional encoding, segmentation size, and finally class query embedding. Our results suggest that although incorporating additional information is beneficial, the most minimal graph still captures sufficient information from the surgical scene enabling comparable phase recognition. This validates our assumption that surgical scene graphs can provide a minimal yet holistic representation of the scene and effectively be used for downstream tasks.

\begin{table*}[]
    \centering
    \caption{Experiments on static and dynamic scene graph construction for phase segmentation of CATARACTS. Mask2Former abbreviated as M2F. Node Features: Spatial Position, Segmentation Size, M2F Embedding, Temporal Position}
    \begin{tabular}{|c|c|cccc|cc|}
          \toprule
          & Segmentation & \multicolumn{4}{|c|}{Node Features} &   &  \\
         Graph  & Backbone & Spatial & Size & Emb & Temp &  Accuracy & $F_1$ \\
         \midrule
         Static & M2F & & & & &  64.34 &  50.04  \\
         Static & M2F & \checkmark & & & & 65.36 & 51.08 \\
         Static & M2F & & \checkmark & & & 64.92 & 48.91 \\
         Static & M2F & & &\checkmark & & 64.47 & 49.05 \\
         Static & M2F &\checkmark & \checkmark & & & 65.56 & 52.24 \\
         Static & M2F &\checkmark &  & \checkmark& & 66.00 & 52.62 \\
         Static & M2F & & \checkmark & \checkmark& & 66.04 & 52.82 \\
         Static & M2F &\checkmark& \checkmark & \checkmark& & \textbf{66.42} & \textbf{53.90} \\
        \midrule
        Dynamic & OCRNet &\checkmark  & \checkmark & & & 54.02 & 39.64 \\
        Dynamic & OCRNet &\checkmark  & \checkmark & &\checkmark & 71.99 & 63.51 \\
        Dynamic & M2F &\checkmark & \checkmark & & & 60.05 & 48.85\\
        Dynamic & M2F &\checkmark & \checkmark & & \checkmark & 73.77 & 64.93\\
        Dynamic & M2F &\checkmark  & \checkmark &\checkmark & & 58.26 & 47.41 \\
        Dynamic & M2F &\checkmark  & \checkmark &\checkmark & \checkmark & \textbf{75.15} &  \textbf{68.56} \\
         \bottomrule
    \end{tabular} \label{tab:ablation_architecture}
\end{table*}

\begin{figure*} [h]
    \centering
    \includegraphics[width=\textwidth]{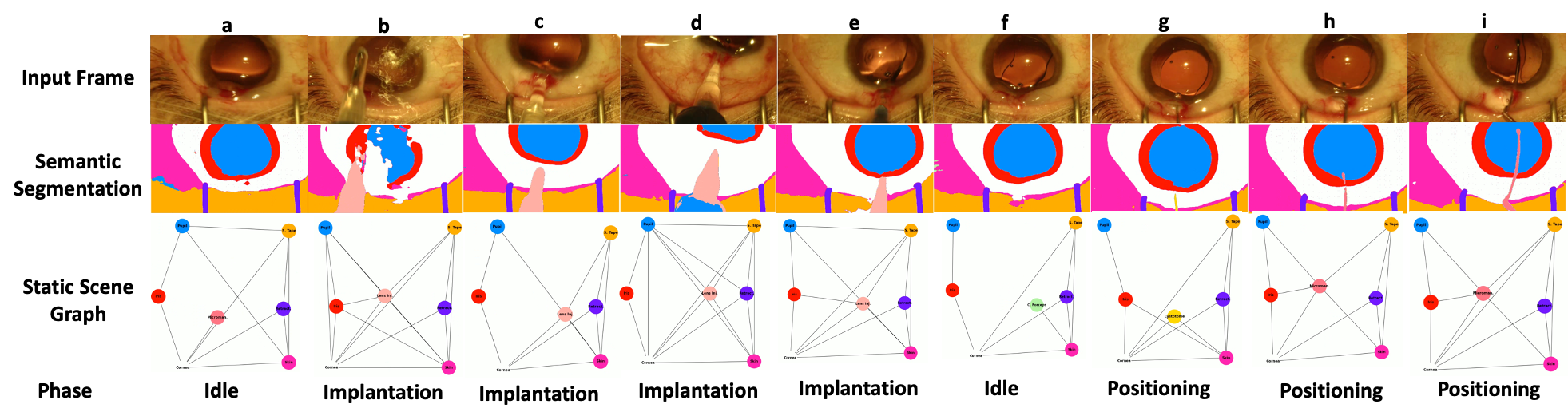}
    \caption{A demonstration of robustness of static scene graphs and more specifically prediction of downstream task of phase segmenatation to noisy semantic segmentation of frames both in localization (b) and classification (g) of objects.}
    \label{fig:robutness}
\end{figure*}

\paragraph{Robustness}
We constructed our scene graphs based on the output of both OCRNet~\cite{pissas2021effective} and Mask2Former~\cite{cheng2022masked} semantic segmentation maps (see Table~\ref{tab:ablation_architecture}). We show that both methods can effectively learn the downstream task, although both backbones provide imperfect segmentation results (see Table~\ref{tab:semantic_seg} and Figure~\ref{fig:robutness}), which demonstrates the inherent robustness of the scene graph representation. As Figure~\ref{fig:robutness} indicates, the phase segmentation results remain consistent along the frames $a$ to $i$ although the semantic segmentation results from Mask2Former indicate incosistencies in this scenario. It can be observed that parts of pupil, iris and lense injector are missed in frame~\ref{fig:robutness}~b and the wrong instrument category is predicted in frame~\ref{fig:robutness}~g. Furthermore, the static scene graph representation indicates a stable scene representation not sensitive to imperfections of semantic segmentation maps which is essential for a reliable surgical workflow analysis.

\paragraph{Temporality}
Table~\ref{tab:ablation_temporal} shows the impact of the length of temporal sequences and learned context via temporal dilation. Our results suggest that a temporal context of $90$s is optimal for classifying phases within Cataract surgery. Considering the duration of phases generally between $5$ s and $2$ min respectively, this temporal context seems appropriate.

\begin{table}
    \centering
    \caption{Experiments on size and stride for the temporal window.}
    \begin{tabular}{ccc|cc}
          \toprule
         Window  & Dilation & Context & Accuracy & $F_1$ \\
         \midrule
         30 & 1 & 30s&   71.72 &  62.61  \\
         30 & 3 & 90s&  \textbf{75.15} &  \textbf{68.56}  \\
         30 & 5 & 150s& 73.24 &  65.67 \\
         30 & 10 & 300s& 73.99 &  65.27  \\
         100 & 1 & 100s& 74.83 &  68.27  \\
         100 & 3 & 300s& 73.15 &  64.75  \\
         100 & 5 & 500s& 71.24 &  62.44  \\
         150 & 1 & 150s& 73.98 &  66.56  \\
         150 & 3 & 450s& 70.32 &  63.93  \\
         150 & 5 & 750s& 69.01 &  60.88  \\
         \bottomrule
    \end{tabular} \label{tab:ablation_temporal}
\end{table}

\paragraph{Performance}
As there are no published benchmarks available on the CATARACTS phase segmentation dataset as publicly released, we provide a comparison to DeepPhase~\cite{zisimopoulos2018deepphase}. DeepPhase is based on the same videos of CATARACTS but uses a different set of labels with 14 phases, from which some are not included in our ground truth and vice-versa. The results we can show here (Table~\ref{tab:deepphase}) indicate that our scene graph-based model performs competitively. 

\begin{table*}[]
    \centering
    \caption{Comparison to DeepPhase~\cite{zisimopoulos2018deepphase} on phase recognition on CATARACTS.}
    \begin{tabular}{lcccc}
         Temporality & Model & Input to temporal Model & Accuracy & $F_1$ \\
         \toprule
         LSTM & DeepPhase~\cite{zisimopoulos2018deepphase} & Binary &  68.75 &  68.50 \\
         LSTM & DeepPhase~\cite{zisimopoulos2018deepphase} & Features &  78.28 &  74.92 \\
         GRU & DeepPhase~\cite{zisimopoulos2018deepphase} & Binary &  71.61 &  67.33 \\
         GRU & DeepPhase~\cite{zisimopoulos2018deepphase} & Features &  68.96 &  66.62 \\
         \midrule
         \midrule
         Dynamic Graph & M2F + GCN & SG + Embeddings & 76.86 & 67.41 \\
         \bottomrule
    \end{tabular}
    \label{tab:deepphase}

\end{table*}

\begin{figure*}[!htbp]
    \centering
    \includegraphics[width=\textwidth]{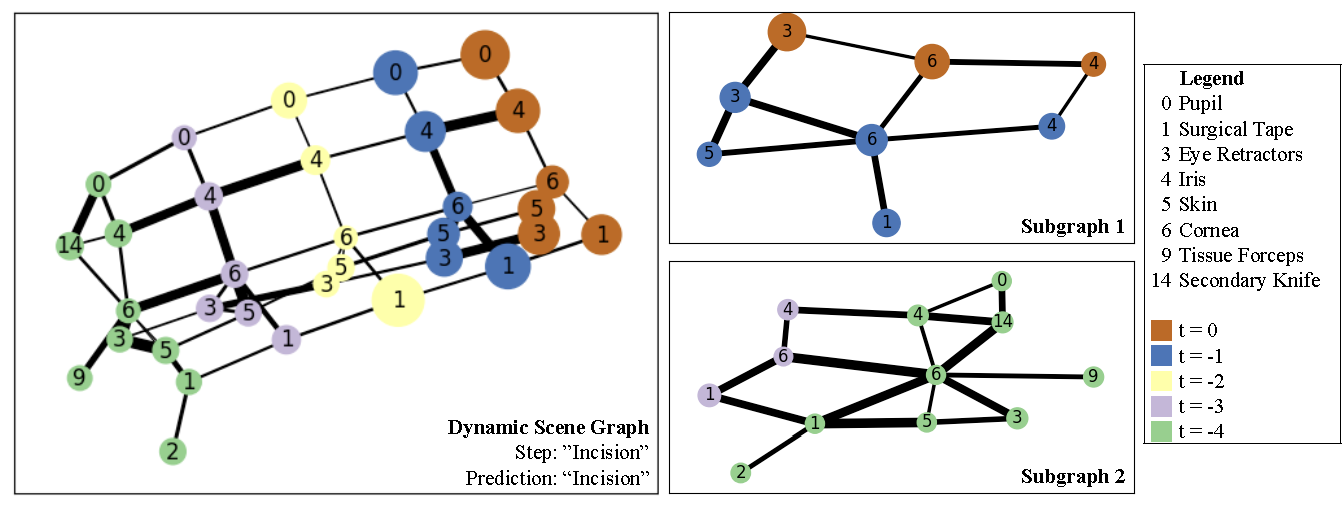}
    \caption{Explanations for Predictions based on the DSG generated with~\cite{ying2019gnnexplainer}. Edge Thickness corresponding to Edge Importance, Node Size corresponding to Node Importance. The Dynamic Scene Graph shows five timesteps, indicated by color. Two important subgraphs can be identified, which are shown on the right.}
    \label{fig:explanation}
\end{figure*}

\paragraph{Explainability}
Finally, we highlight that scene graphs can provide additional insights into the model decisions via direct access to the nodes and the relations between them. Figure~\ref{fig:explanation} highlights the relations and nodes in the DSG that were most important for the prediction. Because of the nature of the scene graphs, these explanations are directly semantically meaningful. E.g., the interaction between the Secondary Knife (14) and Pupil (0), and Tissue Forceps (9) and and Cornea (6) are important for the prediction, even when they occured 5 timesteps in the past. Other realtions that represent temporal connections are highlighted as important as well, such as the relation between the Iris (4) in timestep 0 and -1. This indicates that the changes over time between these timesteps were important for the prediction. From the edge and node importance, two subgraphs can be identified that are shown in Figure~\ref{fig:explanation} on the right. Subgraph 1 consists mostly of anatomy such as Iris, Pupil and Cornea and their changes over the last timestep. Subgraph 2 shows the tool interaction with Secondary Knife and Tissue Forceps and its relation with the surrounding anatomy. This shows our model considers both these spatially and temporally defined regions of the DSG as important for the final correct prediction of step "Incision". We consider the ability for explanation of predictions a major benefit of DSG over latent vector representations, which is crucial in the medical setting to gain the trust of clinicians and patients and ensure their safety through quality control.

\section{Conclusion}
We demonstrate a new method to leverage surgical scene graphs for surgical workflow recognition in cataract surgery. To achieve this, we generate a new scene graph dataset from semantic segmentations based on the CaDIS dataset. We develop a dynamic GCN-based framework to leverage scene graphs for surgical workflow understanding by learning a minimal yet holistic intermediate representation of surgical videos. Our approach proves to be robust, as we apply it to various imperfect segmentation backbones. Furthermore, we provide visualizations of the model's reasoning, highlighting the most important elements of the scene graph, which shows the human understandability and explainability of our approach. We thereby demonstrate the benefits of scene graphs as a representation of surgical video in future surgical data science applications.

\clearpage
{\small
\bibliographystyle{ieee_fullname}
\bibliography{ref}

\begin{thebibliography}{10}\itemsep=-1pt

\bibitem{mask2formercode}
Mask2former.
\newblock \url{https://huggingface.co/docs/transformers/model_doc/mask2former},
  2022.
\newblock Accessed: 2023-02-24.

\bibitem{al2019cataracts}
Hassan Al~Hajj, Mathieu Lamard, Pierre-Henri Conze, Soumali Roychowdhury,
  Xiaowei Hu, Gabija Mar{\v{s}}alkait{\.e}, Odysseas Zisimopoulos, Muneer~Ahmad
  Dedmari, Fenqiang Zhao, Jonas Prellberg, et~al.
\newblock Cataracts: Challenge on automatic tool annotation for cataract
  surgery.
\newblock {\em Medical image analysis}, 52:24--41, 2019.

\bibitem{ac97-8m18-21}
Hassan ALHAJJ, Mathieu Lamard, Pierre-henri Conze, Béatrice Cochener, and
  Gwenolé Quellec.
\newblock Cataracts, 2021.

\bibitem{cheng2022masked}
Bowen Cheng, Ishan Misra, Alexander~G Schwing, Alexander Kirillov, and Rohit
  Girdhar.
\newblock Masked-attention mask transformer for universal image segmentation.
\newblock In {\em Proceedings of the IEEE/CVF Conference on Computer Vision and
  Pattern Recognition}, pages 1290--1299, 2022.

\bibitem{martel_tecno_2020}
Tobias Czempiel, Magdalini Paschali, Matthias Keicher, Walter Simson, Hubertus
  Feussner, Seong~Tae Kim, and Nassir Navab.
\newblock {TeCNO}: {Surgical} {Phase} {Recognition} with {Multi}-stage
  {Temporal} {Convolutional} {Networks}.
\newblock In Anne~L. Martel, Purang Abolmaesumi, Danail Stoyanov, Diana Mateus,
  Maria~A. Zuluaga, S.~Kevin Zhou, Daniel Racoceanu, and Leo Joskowicz,
  editors, {\em Medical {Image} {Computing} and {Computer} {Assisted}
  {Intervention} – {MICCAI} 2020}, volume 12263, pages 343--352. Springer
  International Publishing, Cham, 2020.
\newblock Series Title: Lecture Notes in Computer Science.

\bibitem{czempiel_opera_2021}
Tobias Czempiel, Magdalini Paschali, Daniel Ostler, Seong~Tae Kim, Benjamin
  Busam, and Nassir Navab.
\newblock {OperA}: {Attention}-{Regularized} {Transformers} for {Surgical}
  {Phase} {Recognition}.
\newblock {\em arXiv:2103.03873 [cs]}, Mar. 2021.
\newblock arXiv: 2103.03873.

\bibitem{DBLP:journals/corr/abs-2103-09712}
Xiaojie Gao, Yueming Jin, Yong{-}Hao Long, Qi Dou, and Pheng{-}Ann Heng.
\newblock Trans-svnet: Accurate phase recognition from surgical videos via
  hybrid embedding aggregation transformer.
\newblock {\em CoRR}, abs/2103.09712, 2021.

\bibitem{Garrow2020}
Carly Garrow, Karl-Friedrich Kowalewski, Linhong Li, Martin Wagner, Mona
  Schmidt, Sandy Engelhardt, Daniel Hashimoto, Hannes Kenngott, Sebastian
  Bodenstedt, Stefanie Speidel, Beat Müller, and Felix Nickel.
\newblock Machine learning for surgical phase recognition a systematic review.
\newblock {\em Annals of Surgery}, Publish Ahead of Print, 11 2020.

\bibitem{grammatikopoulou2019cadis}
Maria Grammatikopoulou, Evangello Flouty, Abdolrahim Kadkhodamohammadi,
  Gwenol'e Quellec, Andre Chow, Jean Nehme, Imanol Luengo, and Danail Stoyanov.
\newblock Cadis: Cataract dataset for image segmentation.
\newblock {\em arXiv preprint arXiv:1906.11586}, 2019.

\bibitem{islam_learning_2020}
Mobarakol Islam, Lalithkumar Seenivasan, Lim~Chwee Ming, and Hongliang Ren.
\newblock Learning and {Reasoning} with the {Graph} {Structure}
  {Representation} in {Robotic} {Surgery}.
\newblock In Anne~L. Martel, Purang Abolmaesumi, Danail Stoyanov, Diana Mateus,
  Maria~A. Zuluaga, S.~Kevin Zhou, Daniel Racoceanu, and Leo Joskowicz,
  editors, {\em Medical {Image} {Computing} and {Computer} {Assisted}
  {Intervention} – {MICCAI} 2020}, Lecture {Notes} in {Computer} {Science},
  pages 627--636, Cham, 2020. Springer International Publishing.

\bibitem{DBLP:journals/corr/abs-1912-06992}
Jingwei Ji, Ranjay Krishna, Li Fei{-}Fei, and Juan~Carlos Niebles.
\newblock Action genome: Actions as composition of spatio-temporal scene
  graphs.
\newblock {\em CoRR}, abs/1912.06992, 2019.

\bibitem{kipf2016semi}
Thomas~N Kipf and Max Welling.
\newblock Semi-supervised classification with graph convolutional networks.
\newblock {\em arXiv preprint arXiv:1609.02907}, 2016.

\bibitem{DBLP:journals/corr/KrishnaZGJHKCKL16}
Ranjay Krishna, Yuke Zhu, Oliver Groth, Justin Johnson, Kenji Hata, Joshua
  Kravitz, Stephanie Chen, Yannis Kalantidis, Li{-}Jia Li, David~A. Shamma,
  Michael~S. Bernstein, and Li Fei{-}Fei.
\newblock Visual genome: Connecting language and vision using crowdsourced
  dense image annotations.
\newblock {\em CoRR}, abs/1602.07332, 2016.

\bibitem{pissas2021effective}
Theodoros Pissas, Claudio~S Ravasio, Lyndon Da~Cruz, and Christos Bergeles.
\newblock Effective semantic segmentation in cataract surgery: What matters
  most?
\newblock In {\em Medical Image Computing and Computer Assisted
  Intervention--MICCAI 2021: 24th International Conference, Strasbourg, France,
  September 27--October 1, 2021, Proceedings, Part IV 24}, pages 509--518.
  Springer, 2021.

\bibitem{sharma_rendezvous_2022}
Saurav Sharma, Chinedu~Innocent Nwoye, Didier Mutter, and Nicolas Padoy.
\newblock Rendezvous in {Time}: {An} {Attention}-based {Temporal} {Fusion}
  approach for {Surgical} {Triplet} {Recognition}, Nov. 2022.
\newblock arXiv:2211.16963 [cs].

\bibitem{twinanda_endonet_2016}
Andru~P. Twinanda, Sherif Shehata, Didier Mutter, Jacques Marescaux, Michel de
  Mathelin, and Nicolas Padoy.
\newblock {EndoNet}: {A} {Deep} {Architecture} for {Recognition} {Tasks} on
  {Laparoscopic} {Videos}.
\newblock {\em arXiv:1602.03012 [cs]}, May 2016.
\newblock arXiv: 1602.03012.

\bibitem{DBLP:journals/corr/XuZCF17}
Danfei Xu, Yuke Zhu, Christopher~B. Choy, and Li Fei{-}Fei.
\newblock Scene graph generation by iterative message passing.
\newblock {\em CoRR}, abs/1701.02426, 2017.

\bibitem{ying2019gnnexplainer}
Zhitao Ying, Dylan Bourgeois, Jiaxuan You, Marinka Zitnik, and Jure Leskovec.
\newblock Gnnexplainer: Generating explanations for graph neural networks.
\newblock {\em Advances in neural information processing systems}, 32, 2019.

\bibitem{zisimopoulos2018deepphase}
Odysseas Zisimopoulos, Evangello Flouty, Imanol Luengo, Petros Giataganas, Jean
  Nehme, Andre Chow, and Danail Stoyanov.
\newblock Deepphase: surgical phase recognition in cataracts videos.
\newblock In {\em Medical Image Computing and Computer Assisted
  Intervention--MICCAI 2018: 21st International Conference, Granada, Spain,
  September 16-20, 2018, Proceedings, Part IV 11}, pages 265--272. Springer,
  2018.

\bibitem{ozsoy_4d-or_2022}
Ege Özsoy, Evin~Pınar Örnek, Ulrich Eck, Tobias Czempiel, Federico Tombari,
  and Nassir Navab.
\newblock 4d-or: Semantic scene graphs for or domain modeling.
\newblock In {\em International Conference on Medical Image Computing and
  Computer-Assisted Intervention}. Springer, 2022.

\end{thebibliography}
}

\end{document}